\documentclass[runningheads]{llncs}
\usepackage[T1]{fontenc}
\usepackage{graphicx}
\usepackage{xcolor}
\usepackage{amsmath}
\usepackage{amssymb}
\usepackage{booktabs}
\usepackage{multirow}
\usepackage{colortbl}
\usepackage{algorithm}
\usepackage{algcompatible}
\algnewcommand\algorithmicinitialize{ \textbf{Initialize:}}
\algnewcommand\INIT{\item[\algorithmicinitialize]}%
\newcommand{\yr}[1]{{\color{black}{#1}}}
\begin{document}
\title{Source-Free Test-Time Adaptation For Online Surface-Defect Detection}
\author{Yiran Song \orcidID{0009-0003-6619-7889} \and
Qianyu Zhou\orcidID{0000-0002-5331-050X} \and \\
Lizhuang Ma\orcidID{0000-0003-1653-4341}}
\authorrunning{Yiran. S, Qianyu. Z et al.}
\institute{Shanghai Jiao Tong University \\
\email{\{songyiran,zhouqianyu,lzma\}@sjtu.edu.cn}\\}
\maketitle              % typeset the header of the contribution
\begin{abstract}
Surface defect detection is significant in industrial production. However, detecting defects with varying textures and anomaly classes during the test time is challenging. This arises due to the differences in data distributions between source and target domains. Collecting and annotating new data from the target domain and retraining the model is time-consuming and costly. In this paper, we propose a novel test-time adaptation surface-defect detection approach that adapts pre-trained models to new domains and classes during inference. Our approach involves two core ideas. Firstly, we introduce a supervisor to filter samples and select only those with high confidence to update the model. This ensures that the model is not excessively biased by incorrect data. Secondly, we propose the augmented mean prediction to generate robust pseudo labels and a dynamically-balancing loss to facilitate the model in effectively integrating classification and segmentation results to improve surface-defect detection accuracy. Our approach is real-time and does not require additional offline retraining. Experiments demonstrate it outperforms state-of-the-art techniques. 

\end{abstract}
\section{Introduction}
With the advent of deep learning~\cite{he2016deep,liang2024survey,duan2024mutexmatch,duan2022rda,duan2023towards,zhou2023transvod,he2021end,zhou2024ppr,zhou2024adv2,liu2024emphasizing,wang2024continuous,yang2023zdl}, surface defect detection (SDD)~\cite{CADN} has made great progress recently in detecting surface defects of industrial scenarios. Unfortunately, gathering and labeling anomalous samples is costly. The collected datasets are usually limited, which hinders effective training. As a result, models excel under the same training distribution but suffer from accuracy degradation due to domain shifts, \emph{e.g.,}, varying textures, and new defect classes, which usually appear in testing. 

Test-time Adaptation (TTA) is a task that uses unsupervised testing data to infer the target domain distribution. The online, unlabeled data arrives continuously, demanding immediate model updates and decisions. Various TTA networks have been proposed, such as TENT \cite{wang2020tent} and CoTTA \cite{wang2022continual}. These methods enable models to adapt to different data distributions during the test time. However, directly applying TTA methods to industrial scenarios will encounter several challenges. Firstly, Table \ref{dataset compare} shows that the dataset sizes of industrial datasets are usually significantly smaller compared to classical datasets. A small dataset can lead to a higher likelihood of encountering untrained knowledge and lead the undesirable performance during inference. Besides, different from existing TTA that usually assumes that the source domain and the target domain share the same label space, a more specific challenge in industrial scenarios is that it will encounter novel classes of defects during the online adaptation.

\begin{table} [htb]
\centering
\caption{\textbf{Comparison of classical test-time adaptation dataset and industrial dataset} We can see that for the same TTA task, the dataset for the traditional image segmentation tasks is much larger than that for the surface-defect detection tasks.} 
\vspace{-3mm}
% \resizebox{\linewidth}{!}{
\begin{tabular}{ccc|ccc}
\toprule
      \begin{tabular}[c]{@{}l@{}}Classical \\ Dataset\end{tabular} &  Class  & \begin{tabular}[c]{@{}l@{}}Total \\ number\end{tabular} & \begin{tabular}[c]{@{}l@{}}Industrial \\ dataset\end{tabular}   &  class  & \begin{tabular}[c]{@{}l@{}}Total \\ number\end{tabular}  \\
\midrule
CIFAR10-C & 15*5 & 750000  & KolektorSDD  & 1 & 399\\
ImageNet-C & 15*5 & 3750000  & DAGM  & 10 & 8050   \\
\bottomrule
\end{tabular}
% }  
\label{dataset compare}
\end{table}

%%%%%%%%%%%%%%%%%%%%%%%%%%%%%%%%%%%%%%%%%%%%%%%%%%%%%%%
\begin{figure}[t]
\begin{center}
% \begin{overpic} 
% [width=\linewidth]
% {example-image-a}
% \end{overpic}
\includegraphics[width=\linewidth]{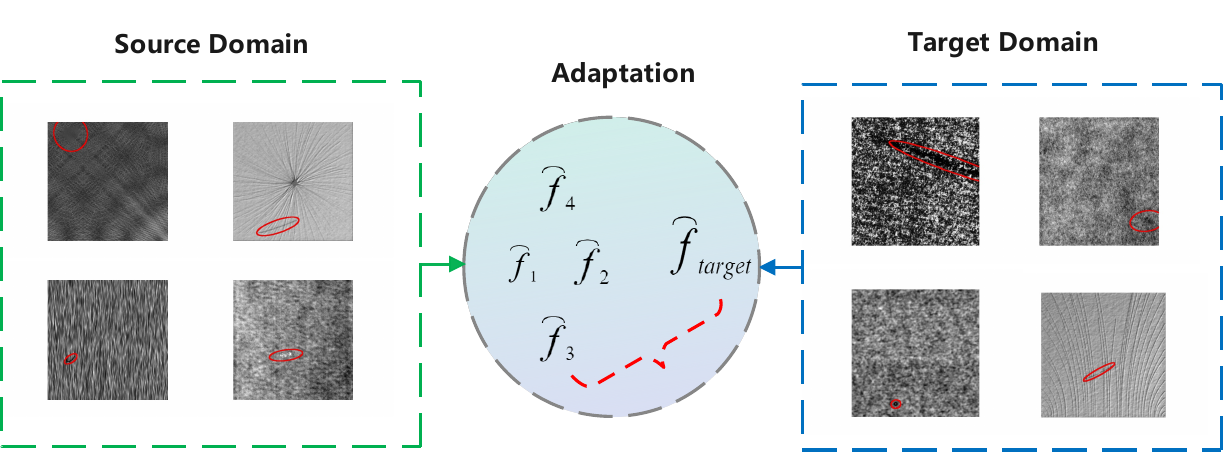}
\end{center}
\vspace{-5mm}
\caption{
Visualization of the domain discrepancy in cross-domain surface defect detection. $f$ represents the optimal parameters that can be learned. Our goal is to find a path that can span the difference between the source and target domains. 
}
\vspace{-3mm}
\label{fig:adaptation}
\end{figure}
%%%%%%%%%%%%%%%%%%%%%%%%%%%%%%%%%%%%%%%%%%%%%%%%%%%%%%%%%%%%%%%%%%%%%%%%%%%%%%%%%%%%55

Motivated by the above analysis, we present a novel test time adaptation method for surface-defect detection. 
\yr{To enhance the adaptability toward the target domains, we introduced a supervisor to predict the sample reliability, which is initialized with source domain parameters, and kept constant during the testing. To bolster the pipeline's robustness, we design two strategies to improve the tranferability: augmented mean prediction and dynamically-balancing loss. Concretely, augmented mean prediction generates multiple predictions per sample and combines them for a more stable pseudo-label. Besides, dynamically-balancing loss adjusts the model's learning focus over time to enhance the robustness of the model. }
Our contributions are summarized as follows. 

$\bullet$ We propose a real-time, test-time adaptation method for online surface-defect detection tasks, without offline retraining or source domain data reuse.

$\bullet$  %To bolster pipeline robustness, 
To bolster pipeline robustness, we introduce a supervisor to filter samples, devise augmented mean prediction, and dynamically-balancing loss to generate more stable pseudo-labels and combat catastrophic forgetting. %which enhances model accuracy and robustness while minimizing noise risk.

$\bullet$ Experimental results show our presented approach outperforms existing state-of-the-art methods on various industrial datasets.

\begin{table*} [t!]
\centering
\caption{\textbf{The difference between our proposed test-time adaptation and related adaptation settings.} We compared the differences in the related settings. Our approach requires only unlabeled test data. The test domain is allowed to have different classes from the source domain. Our approach is online updates on the test domain without source domain data and offline retraining.} 
\vspace{-3mm}
\resizebox{\linewidth}{!}{
\begin{tabular}{c|c c c | c c}
\hline
      setting &   source data  &  target data   & \yr{new} class  & train stage  & test stage    \\
\hline
fine-tuning & no  & stationary+labeled & yes  &  yes & no  \\
standard domain adaptation & yes  & stationary & yes  & yes  & yes  \\
standard test-time training & yes  & stationary & yes  & yes(aux task)  & yes  \\
fully test-time adaptation & no  & stationary & no & no(pre-trained)  & yes    \\
continual test-time adaptation & no  & continually changing & no & no(pre-trained)  & yes    \\
\hline
our industrial setting & no  & continually changing & yes & no(pre-trained)  & yes    \\
\hline
\end{tabular}
}  
\label{methods}
\vspace{-8mm}
\end{table*}

%-------------------------------------------------------------------------

%-------------------------------------------------------------------------
\section{Related Work}

\paragraph{Domain adaptation.}
Our work is related to \textbf{unsupervised} domain adaptation(UDA), \textbf{source-free} domain adaptation, and \textbf{test-time adaptation} (TTA). Though Domain Generalization (DG) methods~\cite{long2024dgmamba,jiang2024dgpic,wang2024disentangle,long2024rethink,zhou2024test,zhou2023instance,zhou2022adaptive,long2023diverse} can improve the model's generalizability, they only utilize the seen data in the training stage, which fails in utilizing the information of the target data, thus resulting in unsatisfactory performance on the target domain.
In contrast, UDA methods \cite{wilson2020survey,hoyer2021daformer,song2023rethinking,song2024ba,zhou2023context,zhou2023self,gu2021pit,zhou2022uncertainty,liu2024cloudmix,zhou2022domainb,guo2021label,xu2021semi,feng2022dmt} aim to adapt a model given unsupervised data, which access labeled data from the source domain and unlabeled data from the target domain at the same time. In our setting, source data is not needed during the adaptation time, and the model is adapted using the unlabeled data solely from the target domain. The source-free domain adaptation methods \cite{li2020model,kundu2020universal,liang2020we,song2024simada,zhou2022generative} require no data from the source domain for the adaptation process. However, most of them are deployed in an offline manner and cannot tackle the online streaming data. 
% TTA techniques \cite{wang2020tent,wang2022continual} aim to adjust the model's parameters without modifying its architecture to adapt to the target domain's characteristics. 

\paragraph{Test-time Adaptation.}
\yr{O}ur work is belong to test-time adaptation\cite{wang2020tent,sun2019test} category. \cite{kurmi2021domain,li2020model,prabhu2021sentry} utilize generative models to acquire feature alignment. Test entropy minimization (TENT)\cite{wang2020tent} is proposed to adapt the test data by minimizing the prediction entropy. 
Source hypothesis transfer (SHOT)\cite{liang2020we} utilizes both entropy minimization and a diversity regularizer for adaptation. \cite{mummadi2021test} apply a diversity regularizer combined with an input transformation module to further improve the performance. 
\cite{karani2021test} uses a separate normalization convolutional network to normalize test images. 
\cite{iwasawa2021test} updates the final classification layer during inference time using pseudo-prototypes. \cite{zhou2021training} proposes a regularized entropy minimization procedure at test-time adaptation, which requires approximating density during training time. 
\cite{hu2021mixnorm,li2016revisiting,you2021test} Update the statistics in the Batch Normalization layer using the target data. \cite{hu2021fully,kundu2021generalize} extend test-time adaptation to semantic segmentation. 

\paragraph{Unsupervised learning for surface-defect detection.} 
Unsupervised learning in industry learns features through reconstruction objective \cite{Chen2017b}, adversarial loss \cite{goodfellow2014generative} or self-supervised objective \cite{Croitoru2017}, without the use of annotated data. Although these methods can significantly reduce the cost of acquiring annotated data, they perform significantly worse compared to fully supervised methods.

\section{Method}
\label{method}
We assume that the model is initialized by parameters $\theta$ pre-trained on the source domains. Our goal is to adjust the model in an online manner during the test time. The input $x_{t}$ is provided at time $t$ sequentially, drawn from target distribution $P^t(\boldsymbol{X}) \neq P^s(\boldsymbol{X})$. The parameters of the model $f_{\theta_{t-1}}$ are updated to $f_{\theta_t}$ in time $t$ based on the input $x_{t}$.
Our setup is motivated by the need for surface-defect detection in industrial scenarios. In Table \ref{methods}, we list the differences between our industrial setup and the relevant adaptation setups that already exist to better show the necessity of our work. Our work focuses on source-free, real-time inference while having fewer constraints on the target domain and higher generalization capabilities. Specifically, we allow the test domain to appear novel anomalous classes and different texture information that do not appear in the source domain, rather than just adding additional noise on the same class \cite{wang2020tent,wang2022continual}. Our setting meets two conditions. 1) Data: only using the unlabeled target domain data. 
2) Updating way: the model is updated online and does not require to be retrained offline.

\begin{figure}[t]
\begin{center}
% \begin{overpic} 
% [width=\linewidth]
% {example-image-a}
% \end{overpic}
\includegraphics[width=\linewidth]{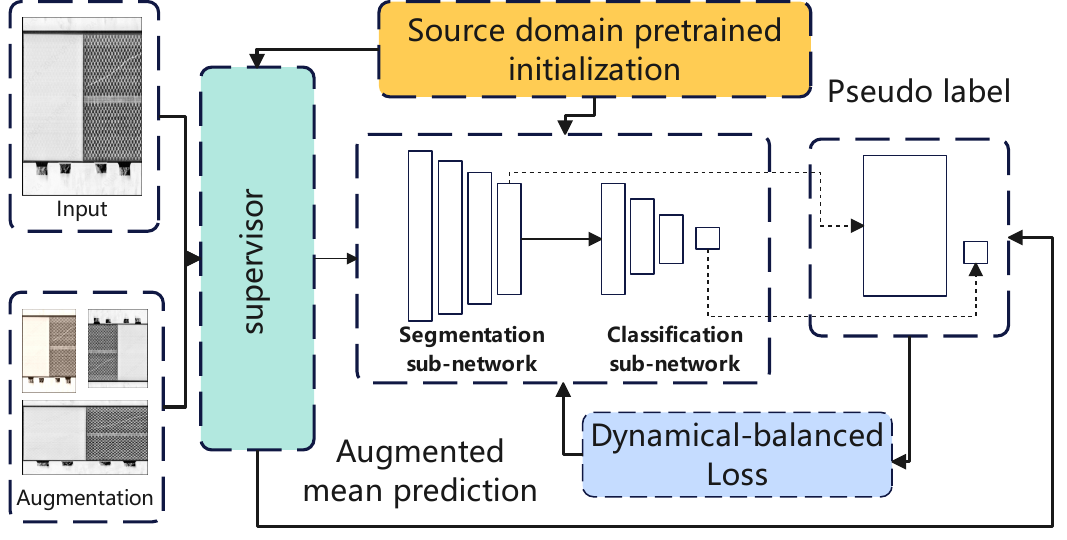}
\end{center}
\vspace{-5mm}
\caption{
\textbf{Architecture of our proposed method -- }
We initialize all modules using the parameters trained on the source domain. Each sample on the target domain is fed to the supervisor to get a score, and only reliable samples are used. The augmented samples are fed into the \textit{supervisor} to obtain prediction results, which are combined with the results inferred from the model to generate the pseudo label. We use the pseudo label to update the model with a \textit{dynamically-balancing loss}.
}
\vspace{-3mm}
\label{fig:pipeline}
\end{figure}

As shown in Figure \ref{fig:pipeline}, our method contains two key parts. Firstly, \textbf{we introduce a supervisor as a ``gate'' to filter the testing data.} We identify untrustworthy data that the model cannot confidently classify, only performing inference on them but excluding them for model updates. For plausible data where the model makes confident predictions, we use them to learn the target data and update the model accordingly. Secondly, we enhance TTA's robustness with two core modules: \textbf{improved average prediction} and \textbf{dynamically balancing loss}. We employ the model's average predictions to reduce outlier impact and boost performance. We also adjust weights in the loss function based on prediction errors to prevent overfitting to training data. 

%%%%%%%%%%%%%%%%%%%%%%%%%%%%%%%%%%%%%%%%%%%%%%%%%%%%%%%
%%%%%%%%%%%%%%%%%%%%%%%%%%%%%%%%%%%%%%%%%%%%%%%%%%%%%%%%%%%%%%%%%%%%%%%%%%%%%%%%%%%%
%-------------------------------------------------------------------------
\paragraph{Base Model.}
We use a lightweight two-stage CNN-architecture model \cite{bovzivc2021end} as our base model. This model is a two-stage end-to-end structure that supervises both the segmentation and classification results during the training time. The number of parameters is much smaller than common network models (\emph{e.g.,} ResNet, ViT, etc.), meeting the performance requirements of industrial detection (i.e., real-time inference updates).

%-------------------------------------------------------------------------
\subsection{Supervisor}
\label{supervisor}

In the test time adaptation testing, using untrustworthy predicted results as pseudo-labels for self-supervised learning can lead to model performance bottlenecks. In industrial scenarios, models are more sensitive to such inaccurate pseudo-labels, because the segmentation results for anomaly detection in industrial settings are usually small, and be more sensitive to subtle changes. Additionally, due to performance constraints, industrial detection models are typically smaller, and using erroneous labels for model updates can cause the model to move even further in the wrong direction. 

To deal with accumulated errors, we create a supervisor with a structure similar to the model. The key distinction is that the supervisor using parameters from source domain training and \textbf{does not} undergo backpropagation. It retains the original knowledge. When the supervisor finds low prediction reliability \yr{$p$}  below a set threshold \yr{$p_{th}$}, the adaptive phase is skipped. This is because pseudo-labels from low-confidence predictions can mislead the model. We skip such samples to avoid steering the model in the wrong direction.

Adapting to new distributions can lead to losing knowledge from the source domain, causing severe information loss. Unlike others, we are constrained to not retrain the model from the source, and prolonged self-training may introduce errors, affecting label accuracy. In our approach, the supervisory module employs pre-trained parameters from the source domain, without further updates in the entire TTA process. It holds source domain knowledge and guides pseudo-label generation, preventing memory loss. 

\subsection{Augmented Mean Prediction}
As shown in section \ref{supervisor}, the supervisor we proposed is initialized with parameters pre-trained on the source domain and is not updated during the test time. It comes with all the knowledge learned from the source domain. We use it to generate pseudo-labels to introduce source domain information. Besides, we propose a method based on augmented average predictions. Specifically, we use sample images that have been data enhanced in many different ways \emph{e.g.,} stretched, cropped, flipped, \emph{etc.}) to input into the supervisor to obtain the prediction results. At the same time, the complete original sample images are also input into the model to obtain $Y$. When performing the filtering operation, each sample is given a confidence value $p$ for the pseudo label. the pseudo label of this image is finally obtained from all the above predictions by weighting averaged. The weight $w$ is a function related to the confidence level $p$ value. The lower the confidence level, \yr{the more the model will refer to the prediction results the supervisor gave} (i.e., source domain knowledge) using the augmented picture. \yr{The specific calculation formula is shown below:
\begin{equation}
\tilde{y}_t  =\frac{1}{N} \sum_{i=0}^{N-1} f_{\theta_t}\left(\operatorname{aug}_i\left(x_t\right)\right) 
\end{equation}
where $p_{{aug}_i} > p_{th}$. Here, $p_{{aug}i}$ and $p{th}$ refer to the confidence of the augmented image and the confidence threshold, respectively.}

\subsection{Dynamically-balancing loss}
Traditional TTA methods compute the loss function based solely on the segmentation or classification results. However, these are not suitable for surface-defect detection. This is because the segmentation portion of anomaly detection datasets is much smaller than the background and the textures are complex, making segmentation quite difficult. In addition, for anomaly detection tasks, correct classification (whether samples with surface defects can be identified) is of great practical significance. Therefore, during the TTA phase, we simultaneously compute the classification and segmentation losses. 

We propose a dynamic weight loss function rather than a fixed weight loss. Specifically, our loss function is defined as:
$\begin{array}{l}L_{\text{total}}=\lambda_{\text{class}}L_{\text{class }}+\left(1-\lambda_{class}\right)L_{\text{seg}}\end{array}\label{loss}$, 
$\lambda_{\text {class }}=\yr{1-t/N}$, \yr{where $t$ is the current-time index and $N$ is the number of test dataset.}  
It utilizes a time-dependent weighting scheme to balance the classification and segmentation losses during the self-adaptive testing phase. The underlying principle is that the model's performance in different tasks changes as it adapts to the target domain. By giving priority to the classification loss at the beginning, the model can focus on correctly classified samples, which is crucial for identifying anomalous regions. As the model's distribution shifts towards the target domain, the segmentation loss is given more weight, enabling the model to capture the complex features of the anomalous regions more accurately.

For the specific calculation of these two components, we tried various combinations of common loss functions, including Kullback–Leibler divergence loss, BCE loss, softmax loss, and DICE loss. \yr{Through our experiments, we find that} the softmax loss combined with Kullback–Leibler divergence loss achieves the best learning effect, specifically defined as follows.
\begin{equation}
\mathrm{L}_{\mathrm{class}}=\frac{1}{n} \sum_{i=1}^n\left(-\log \frac{e^{l_{i, Y(i)}}}{\sum_{k=1}^C e^{l_{i, k}}}\right)
\end{equation}
\begin{equation}
L_{seg}= -\sum x\log(p)- \left(-\sum x\log(x)\right)
\end{equation}

To illustrate our algorithm more clearly, the complete process is shown in Algorithm \ref{alg:Framwork}. Through the filtering by the supervisor and the optimization, our algorithm effectively reduces error accumulation and catastrophic forgetting when test time adaptation is performed on the target domain.

\subsection{Model Update Pipeline}
To illustrate our algorithm more clearly, the complete process is shown in Algorithm \ref{alg:Framwork}. Our algorithm effectively reduces error accumulation and catastrophic forgetting when unsupervised learning is performed on the target domain.
\begin{algorithm}[htb]
\caption{ Framework for online test-time adaptation}
\label{alg:Framwork}
\begin{algorithmic}[1] %这个1 表示每一行都显示数字
\INIT A model $f_{\theta_0}(x)$ and Its supervisor $m_{\theta_0}(x)$ \yr{(Both initialized with parameters $\theta_0 $ which obtained by pre-training on the source domain $D_{s}$). The threshold used for filtering samples $p_th$ }.
\REQUIRE For each time step t, unlabeled data $x_t$ sampled from target domain $D_{t}$ .  \\ %算法的输入参数：Input
    Provide input $x_t$ to the supervisor $m_{\theta_0}(x)$ and obtain the confidence probability $p$;
    \IF{\yr{$p > p_{th}$}}
    \STATE Provide the set of Augment $x_t$ to supervisor $m_{\theta_0}(x)$ and obtain predictions;
    \STATE Provide $x_t$ to model $f_{\theta_0}(x)$ and obtain predictions;
    \STATE Use augmented mean prediction method to acquire pseudo-label of $x_t$
    \STATE Upgrade model $f_{\theta_0}(x)$ by loss in \ref{loss}
    \ENDIF
    \STATE Calculation of prediction result \yr{$y_t$}
\ENSURE Prediction \yr{$y_t$}, Updated model $f_{\theta_t}(x)$
\end{algorithmic}
\end{algorithm}

%------------------------------------------------------------------------
\section{Experiments}

%-------------------------------------------------------------------------
\subsection{Datasets and Pre-training}

\paragraph{DAGM 2007 dataset.}
DAGM dataset\cite{Weimer2016} is a well-known benchmark database for surface-defect detection. It contains images of various surfaces with artificially generated defects. Surfaces and defects are split into 10 classes of various difficulties. We randomly selected four types of samples from the DAGM dataset as the training set for pre-training the model. The model is then no longer exposed to the source domain dataset but is validated on the remaining six unseen anomaly classes. As shown in Figure \ref{fig:dagm}, there are significant differences in the distribution of the ten anomaly classes.

\paragraph{KolektorSDD datasets.} 
\cite{Tabernik2019} is annotated by the Kolektor Group. The images were captured in a controlled industrial environment in a real-world case. The dataset consists of 399 images, of which 52 images with visible defects and 347 images without any defects.
The original width is 500 px, and the height is from 1240 to 1270 px. We resize images to 512 x 1408 for training and evaluation. For each item, the defect is only visible in at least one image, while two items have defects on two images, which means there were 52 images where the defects are visible. The remaining 347 images serve as negative examples with non-defective surfaces. Since KolektorSDD does not have the same subclass division as DAGM, we manually divide the anomalous samples into two parts with large morphological differences and use one part for training while validating the other part to demonstrate its adaptive ability on target domains with different distributions.

\paragraph{Pre-training.}
Following the work in \cite{bovzivc2021end}, we use a two-stage model as our base model, where the segmentation is performed in the first stage, followed by a per-image classification in the second stage. We train the network using stochastic gradient descent with no weight decay and no momentum. \yr{We initialize the base model and the Supervisor using the trained parameters $\theta_0$.} Since the dataset suffers from severe positive and negative sample imbalance, we use low sampling of negative samples, and in each training epoch, we select negative samples of the same size as the positive subset of the sample. We also ensured that all negative images were used approximately equally often during the learning process. Our pre-training on the source domain does not require additional measures to improve the generalization ability of the model. For the DAGM dataset, we train 60 epochs with a learning rate of 0.01. batch size = 5. For the KolektorSDD dataset, we train 35 epochs, using a learning rate of 0.5. After training with the source domain data, the model does not need to use the training data again in the subsequent stages and will not be retrained offline again.
%------------------------------------------------------------------------

\subsection{Results}

\paragraph{Inference Time.} As we have highlighted, our model is a lightweight, online-inference cnn-architecture model to meet the requirements of industrial scenarios. \yr{Our proposed method achieves 32 fps on 512 x 512 images (DAGM) and 13 fps on 512 x 1408 images (KolektorSDD). For TTA methods that do not require retraining: CoTTA achieves 25 fps on 512 x 512 images (DAGM) and 9 fps on 512 x 1408 images (KolektorSDD). This latency is due to the need to update both student and teacher models simultaneously (our method only requires updating one). For other methods that require retraining (unsupervised and weakly supervised), they can achieve faster inference speeds during the inference stage. However, the additional training takes approximately 15 minutes (KSDD) to 30 minutes (DAGM) and results in poorer inference accuracy (as shown in Table 4). By default, all of our results are based on a single Nvidia RTX2080Ti GPU.}

\paragraph{Test Time Setting.} Without special emphasis, we set batch size=1 and use 1e-3 as the learning rate with Adam optimizer. Following \cite{wang2022continual}, we use the same data augmentation operations, including color jitter, random affine, random horizontal flip, and so on. We use 4 augmentations for our experiments. The threshold $p$ is 0.6 by default.

As shown in table \ref{tab:example}, The sample sizes of the surface-defect detection datasets are very small. This makes the base model originally supported by Cotta\cite{wang2022continual} and Tent\cite{wang2020tent} perform poorly on KolektorSDD. (On CIFAR10C\cite{hendrycks2019robustness} they used WideResNet-28\cite{zagoruyko2016wide}, on CIFAR100-C they used ResNeXt-29\cite{xie2017aggregated} and on ImageNet-C\cite{hendrycks2019robustness} they used resnet50\cite{croce2021robustbench}) For a fair comparison, we used the same two-stage model\cite{bovzivc2021end} as the base model, along with the same epoch training parameters to initialize. This allows us to more accurately compare the strengths and weaknesses of the methods in the adaptive phase. It is guaranteed that the difference is not due to a difference in the base model.

\begin{figure}[t]
\begin{center}
\includegraphics[width=0.7\linewidth]{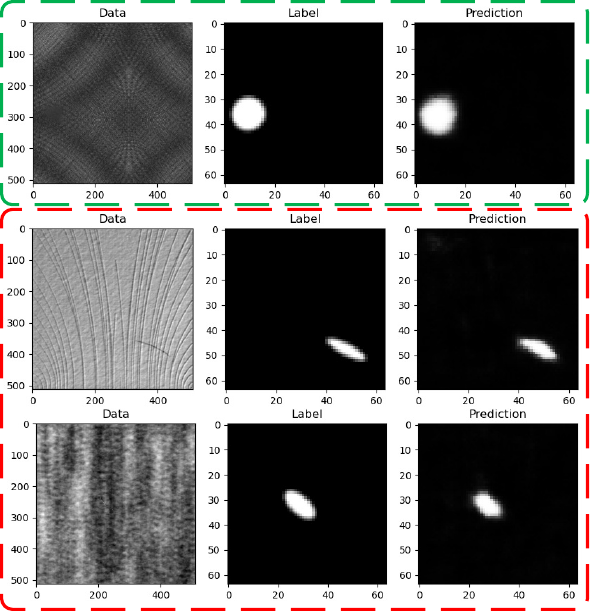}
\end{center}
\vspace{-5mm}
\caption{
\textbf{Visualization of images, labels and segmentations on the DAGM} The green box shows the effect of segmentation within the source domain. The red boxes show the segmentation of the new classes that emerged during the test-time.
}
\vspace{-5mm}
\label{fig:dagm}
\end{figure}

\paragraph{Experiments on DAGM.}
We first validate the effectiveness of our method on the DAGM dataset, which has ten classes, each with different texture and surface anomalies. To verify the reliability and stability of our method, we randomly select four of the ten classes as the source domain for training, while using the remaining six classes as the testing set for testing. This experiment was repeated for several sets. In Table \ref{tab:dagm}, we present the full results comparing the accuracy of inference using our proposed method with inference directly on the testing set, thus demonstrating that our method can improve the inference accuracy of the model on the target domain when the source and target domains do not coincide. Also, as shown in Table \ref{tab:dagm}, for a fair comparison, we compare our method with other TTA methods, demonstrating that our method is more applicable to industrial scenarios. We also compare with unsupervised and weakly supervised methods. For the unsupervised and weakly supervised work, we train on a training set of each class and test on a testing set. For unsupervised and weakly supervised methods, we follow \cite{bovzivc2021mixed}. For TENT \cite{wang2020tent} and CoTTA \cite{wang2022continual}, we use the official open-source code. The results are averaged over all 10 classes. For the TTA work, we use a four-class training set for training and a ten-class test set for inference to demonstrate the model's ability to remember source domain knowledge and adapt to the target domain. We demonstrate that our approach can make better use of the source domain knowledge, combined with the unlabelled target domain knowledge, to obtain better inference accuracy on the target data domain. Figure \ref{fig:dagm} presents some examples of the predictions of our method.

In addition, we have found that TTA methods designed for traditional segmentation tasks do not achieve good accuracy in surface-defect detection. They are even significantly lower than the weakly supervised methods trained on the target domain from scratch. This is mainly due to their complex design (e.g. randomly recovering some of the model parameters as initialization) and their full update (updating the model parameters with every sample) that are not applicable to this task. The high level of instability in the target domain, and the false pseudo-labeling produced, hurt the inference of the model.
%%%%%%%%%%%%%%%%%%%%%%%%%%%%%%%%%%%%%%%%%%%%%%%%%%%%%%%

% \usepackage{multirow}
\begin{table}[ht]
  \centering
\caption{\textbf{Details of the the evaluation datasets.}}
  % \resizebox{\linewidth}{!}{
    \begin{tabular}{lcccp{4.235em}}
    \toprule
    Dataset & \multicolumn{1}{p{4.235em}}{Positive\newline{}Samples} & \multicolumn{1}{p{4.235em}}{Negative\newline{}Samples} & \multicolumn{1}{p{4.235em}}{Defect\newline{}Types} & \multicolumn{1}{l}{Annotations} \\
    \midrule
    DAGM1-6 & 450   & 3000  & 6     & \multicolumn{1}{l}{ellipse} \\
    DAGM7-10 & 600   & 4000  & 4     & \multicolumn{1}{l}{ellipse} \\
    KolektorSDD & 52    & 347   & 1     & bbox \\
    \bottomrule
    \end{tabular}
    % }%
  \label{tab:example}%
\end{table}%

%-------------------------------------------------------------------------

% \usepackage{multirow}
\begin{table}[ht]
\centering
\caption{\textbf{Comparison with state-of-the-art SDD methods on the DAGM dataset.} AP, CA, US, and WS are abbreviations for Average Precision, Classification Accuracy, Unsupervised, and Weakly Supervised, respectively.} 
\resizebox{0.5\linewidth}{!}{
\begin{tabular}{l|c|cccc}
\toprule
Method & \yr{Venue } & Type& CA  & AP  \\ 
\midrule
f-AnoGAN \cite{fAnoGAN}   & \yr{MIA 2019 }          & US                    & 79.7                   & 19.5                   \\
Uninf. stud.\cite{bergmann2020uninformed}  & \yr{CVPR 2020} & US        & 84.3     & 66.8                   \\
Staar  \cite{Staar2018}  & \yr{CIRP 2019} & US    & -      & -      \\ \midrule
CADN-W18\cite{CADN}   &  \yr{PR 2021} &WS     & 86.2       & -     \\
CADN-W18(KD)\cite{CADN}  &  \yr{PR 2021}   & WS    & 87.6     & -      \\
CADN-W32 \cite{CADN} &  \yr{PR 2021}   & WS    & 89.1 & -           \\ \midrule
TNET  \cite{wang2020tent} &  \yr{ICLR 2021}  & TTA    &  86.3   &   85.1  \\
CoTTA \cite{wang2022continual}  &  \yr{CVPR 2022}  & TTA    &   85.2     & 84.4   \\
EATA \cite{niu2022efficient} &  ICML 2022   & TTA    &   89.3     & 90.1   \\
SAR \cite{niu2023towards} &  ICLR 2023   & TTA    &   87.9     & 86.1   \\
DeYO \cite{lee2024entropy}  &  ICLR 2024  & TTA    &  90.4     & 90.6   \\
Our method  & TTA  &  -  &   90.3     &       89.2     \\ 
\bottomrule
\end{tabular}
 }
\label{tab:dagm}
\end{table}

%--------------------------------------------------------------

\begin{table}[ht]
\centering
\caption{\textbf{Comparison with prior works on KolektorSDD dataset.} For unsupervised and weakly supervised methods, we follow the official codes of \cite{bovzivc2021mixed}. For tent\cite{wang2020tent}.}
% \resizebox{0.65\linewidth}{!}{
\begin{tabular}{l|ccccc}
\toprule
Method & f-AnoGAN  & Uninf. stud.  & \cite{bovzivc2021mixed} &  TNET &  \textbf{Our method} \\ 
\midrule
 AP & 39.4  & 57.1  & 93.4  & 92.1 & 94.7 \\
\bottomrule
\end{tabular}
 % }
\label{tab:KolektorSDD}
\end{table}

\begin{figure}[t]
\begin{center}
\includegraphics[width=0.65\linewidth]{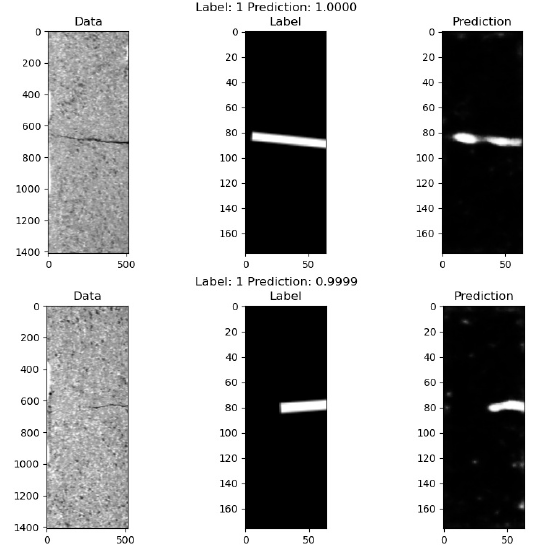}
\end{center}
\vspace{-5mm}
\caption{
\textbf{Examples of predictions from the KolektorSDD}
}
\vspace{-5mm}
\label{fig:KolektorSDD}
\end{figure}

\begin{table}[]
\centering
\caption{Performence of individual components on DAGM dateset}
% \resizebox{0.9\linewidth}{!}{
\begin{tabular}{cccc}
\toprule
AP  & \begin{tabular}[c]{@{}l@{}}supervisor \\ filtering\end{tabular}   & \begin{tabular}[c]{@{}l@{}}Augmented mean \\ prediction\end{tabular}   & \begin{tabular}[c]{@{}l@{}}Dynamic \\ balance loss \end{tabular}  \\ 
\midrule
85.7  &  &  &  \\
87.2  & \checkmark &  &  \\
87.9  & \checkmark & \checkmark &  \\
88.5  & \checkmark & \checkmark & \checkmark \\
\bottomrule
\end{tabular}
% }
  \label{tab:ablation}
\end{table}

\paragraph{Experiments on KolektorSDD.}
We validated the effectiveness of our method on the KolektorSDD dataset. Similar to the work done on DAGM, we compare our method with existing TTA methods\cite{wang2020tent}, unsupervised\cite{fAnoGAN,bergmann2020uninformed} and weakly supervised methods\cite{bovzivc2021mixed} as shown in the Table \ref{tab:KolektorSDD}. The KolektorSDD dataset is simpler compared to the DAGM dataset, so we only performed a comparison of AP accuracy (most methods can achieve very high accuracy in classification accuracy). We visualize the prediction results in Figure \ref{fig:KolektorSDD} to verify the effectiveness.

Although we manually filtered the dataset to partition the KolektorSDD dataset into widely varying source and target domains, this was still not sufficient to demonstrate the migration capability of our method. Therefore, we used four randomly selected categories in DAGM for training and then used this pre-trained model directly for adapting on the KolektorSDD dataset. A competitive accuracy of 0.93 was obtained for our model.

\subsection{Ablation study}
Finally, we evaluated the impact of each component,
named supervisor filtering, augmented mean prediction, and dynamic balance loss. Results are reported in Table \ref{tab:ablation}. We conducted ablation studies on the DAGM and KolektorSDD datasets. We used the same number of samples for testing, uniformly initialized with a pre-trained model trained for 50 epochs. We report performance by progressively enabling individual components and disabling specific components while retaining all remaining components. The results are reported in Table 4. The results show that on all three datasets, the worst performance was achieved with no components enabled, while the best performance was achieved with all three components. Below, we describe in detail the contribution of each component to the overall improvement.

The use of a supervisor to filter the sample data yields the greatest accuracy gain for this method. The large discrepancy between the source and target domain data, along with the small sample data size and small model size, leads to a more pronounced accumulation of errors if incorrect pseudo-labels are used to train the model. Such errors can lead to even greater errors from subsequent pseudo-labels, resulting in worse model accuracy compared to direct inference in the target domain (without test time adaptation). Using a supervisor to filter noisy samples mitigates this problem well.

All our designed components for test time adaptation, including augmented mean prediction and dynamic balance loss, contributed to the accuracy improvement compared to testing directly on the target domain without adaptation. This demonstrates that the method we have designed is effective in improving the robustness of the model.

%-------------------------------------------------------------------------
\section{Conclusion}
In this paper, we propose a novel online test-time adaptation framework for surface-defect detection, which addresses the challenge of detecting unforeseen anomalies in product surfaces during test time. We introduce the parameter-frozen supervisor to allow the model to remember the source domain knowledge over time, while continuously updating the model parameters to adapt to the distribution of the target domain. To bolster pipeline robustness, we devise augmented mean prediction and dynamically-balancing loss.  % We believe that our research direction is of great interest, 
Considering the difficulty and cost of collecting anomalous samples, our framework not only saves time and resources but also enhances the efficiency of the detection process. 
Thus, our method offers a promising solution to the %challenges of 
surface-defect detection in industrial production processes.
Experimental results demonstrate that our method yields superior inference accuracy on both the source and target domains.

\subsubsection{Acknowledgements} Thanks to the support of Shanghai Municipal Science and Technology Major Project (2021SHZDZX0102), Shanghai Science and Technology Commission  (21511101200), National Natural Science Foundation of China (No. 72192821), YuCaiKe [2023] Project Number: 14105167-2023.
% Please place your acknowledgments at
% the end of the paper, preceded by an unnumbered run-in heading (i.e.
% 3rd-level heading).

\bibliographystyle{splncs04}
\bibliography{ref}




% LNCS Homepage, \url{http://www.springer.com/lncs}. Last accessed 4
% \end{thebibliography}
\end{document}